# RAIM: Recurrent Attentive and Intensive Model of Multimodal Patient Monitoring Data

Yanbo Xu
Georgia Institute of Technology
yxu465@gatech.edu

Siddharth Biswal
Georgia Institute of Technology
sbiswal7@gatech.edu

Shriprasad R Deshpande
Emory University School of Medicine
Children's Healthcare of Atlanta
DeshpandeS@kidsheart.com

Kevin O Maher
Emory University School of Medicine
Children's Healthcare of Atlanta
komaher@emory.edu

Jimeng Sun
Georgia Institute of Technology
jsun@cc.gatech.edu

## ABSTRACT

With the improvement of medical data capturing, vast amount of continuous patient monitoring data, e.g., electrocardiogram (ECG), real-time vital signs and medications, become available for clinical decision support at intensive care units (ICUs). However, it becomes increasingly challenging to model such data, due to high density of the monitoring data, heterogeneous data types and the requirement for interpretable models.

Integration of these high-density monitoring data with the discrete clinical events (including diagnosis, medications, labs) is challenging but potentially rewarding since richness and granularity in such multimodal data increase the possibilities for accurate detection of complex problems and predicting outcomes (e.g., length of stay and mortality). We propose Recurrent Attentive and Intensive Model (RAIM) for jointly analyzing continuous monitoring data and discrete clinical events. RAIM introduces an efficient attention mechanism for continuous monitoring data (e.g., ECG), which is guided by discrete clinical events (e.g, medication usage). We apply RAIM in predicting physiological decompensation and length of stay in those critically ill patients at ICU. With evaluations on MIMIC-III Waveform Database Matched Subset, we obtain an AUC-ROC score of 90.18% for predicting decompensation and an accuracy of 86.82% for forecasting length of stay with our final model, which outperforms our six baseline models.

## 1 INTRODUCTION

Electronic health record (EHR) data consist of event sequences such as diagnosis and medication prescriptions, vital signs and lab results. There are many recent deep learning successes in modeling such event datasets.[4, 5, 8, 9, 12, 28]. However, modern healthcare practice, especially inpatient and intensive care, also generates vast amount of heterogeneous monitoring data in real-time such as electrocardiogram (ECG), pulse plethysmograms, and respirations. The availability of rich and massive amount of patient monitoring data opens an opportunity for developing accurate models for better clinical decision support.

In this paper, we are interested in integrating continuous monitoring data with discrete clinical event sequences and developing clinical predictive models in ICUs. The goal is to detect physiological deterioration more precisely in these critically ill patients and also predict their length of stay more accurately. There exist three main challenges in this work:

- **Multi-channel high-density signal processing.** Patients are constantly monitored by multiple special equipments at the bedside in ICUs. Vast amount of high-dimensional streaming data are captured in real-time from each patient and these data become humongous as patients stay at ICUs from days to weeks. For example, a patient staying at ICU for one day can generate up to 11M values from a single lead ECG recording sampled at 125Hz and 86K values per each vital sign sampled minutely. Thus a computationally efficient modeling approach is needed to handle these multi-channel high-density input signals as well as the dynamic temporal behaviors within the sequences.
- **Multiple data modalities.** When jointly modeling dense and sparse data, the dense physiological data can potentially dominate the learnt representations and mask beneficial informations from the sparse clinical event data. Special modeling strategy is needed for handling such variability in data density.
- **Interpretability.** Many machine learning based methods, such as deep neural networks (DNN), are treated as a black box for many application domains. However, an interpretable model is important for clinical decision support applications as the predictive results need to be understood by clinicians.

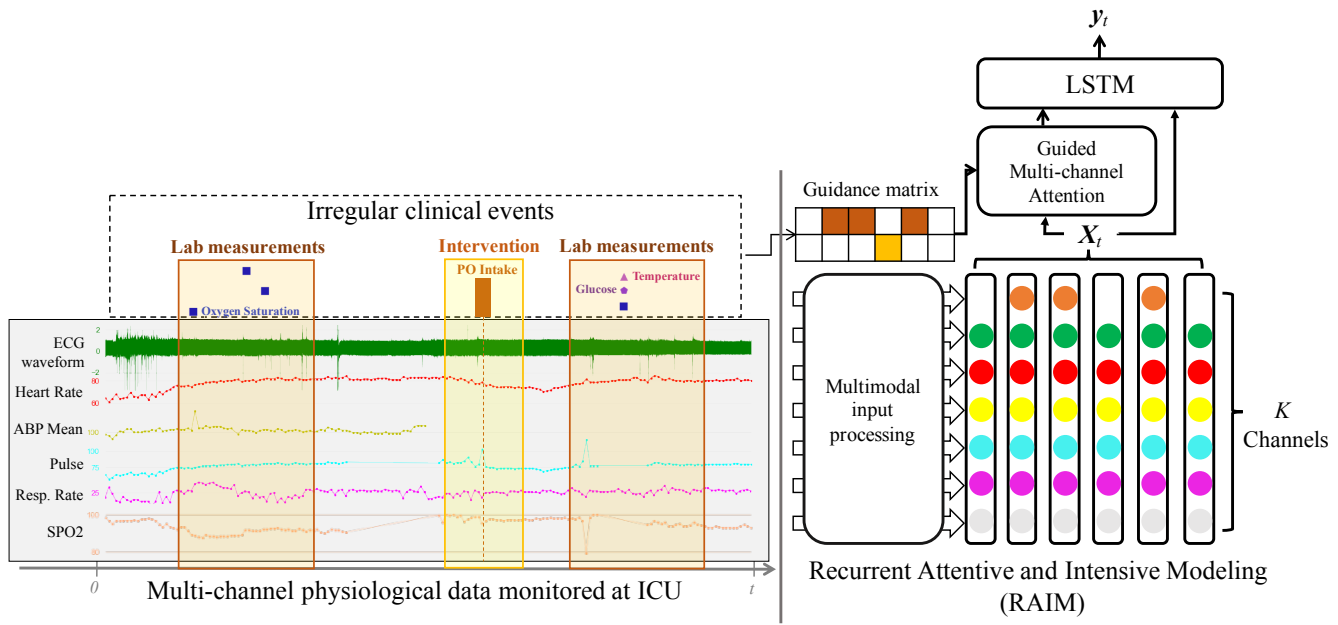

Figure 1: An overview of RAIM on multimodal continuous patient monitoring data.

To address the above challenges, we propose RAIM, a Recurrent Attentive and Intensive Model for analyzing the multimodal EHR time series collected at ICUs. Compared to the conventional attention mechanism in deep networks for single modalities, RAIM proposes an efficient multi-channel attention on continuous monitored data, which is guided by discrete clinical data. Figure 1 gives the overview of RAIM: Multi-channel physiological data, such as 125Hz sampled ECG waveforms and continuously updated (per second) vital signs, are integrated with discrete irregular clinical data including lab measurements, and interventions. RAIM generates a guidance matrix from lab measurements and interventions, use it to guide the multi-channel attention mechanism on processed multimodal input streams, and eventually predict the dynamic outputs such as the time-varying risk scores of decompensation.

RAIM advances state-of-the-art in a number of ways. First, in contrast with past works that focus only on either clinical EHR data or continuous monitoring data, RAIM integrates both and study the multiple data modalities together. Second, compared with the recent deep learning methods focusing on classifying short-term waveforms (e.g. ECG), RAIM models long sequence of multi-channel waveforms and predicts the outputs dynamically. Lastly but not least, unlike most of the existing methods that simply aggregate clinical data as one feature vector RAIM breaks it down to different modes and extract different levels of ingredients contained in the data.

## 2 RELATED WORK

**Deep Learning on EHR discrete data.** As Deep learning techniques are gaining popularity in many domains and the amount of EHR clinical data is growing explosively in the past few years, deep learning approaches have been adapted to the data in various clinical applications. For example, variations of convolutional neural network (CNN) and restricted boltzmann machine (RBM) are applied on structural clinical data (e.g. diagnoses, procedures, medications) to learn vector representations of patients or medical concepts [5, 8, 9, 34]; recurrent neural networks (RNNs) like LSTM and GRU are used to capture the sequential manner of EHR data and predict future disease diagnosis or intervention codes [28, 39], onset of heart disease [4] or kidney transplantation related complications [12]; autoencoders (AEs) and RBMs are fitted on raw discrete codes for discovering phenotypes [26, 35]. A more detailed survey of recent deep learning approaches for EHR analysis can be found here [43]. Deep learning methods have been approved in above works to gain better performance than conventional machine learning methods. In addition to use discrete structural EHR data, other works target on unstructured data like clinical notes, the main purpose of which are to extract structured medical concepts [13, 19, 20, 29, 31].

**Deep learning on patient monitoring data.** Traditional signal processing algorithms were widely used for analyzing patient monitoring data, such as P-QRS-T detection [44] or RR interval/heart rate variability (HRV) extraction[1] for ECG signal processing. Only until recently, sufficiently large amount of continuous monitoring data are captured for modeling. Since traditional methods have difficulties in scaling up computationally and extracting features comprehensively on such vast amount of data, deep learning models start to take over the task. Recent work [40] has successfully used CNN to classify heart arrhythmia classes based on raw 30-second single-lead ECG signals. Another work [41] takes time variations into account and use RNN to detect arrhythmia based on extracted ECG features (e.g. R amplitudes, QRS durations and etc). Other works apply LSTM or CNN with temporal convolutions on the raw vital signs and lab sequences to predict mortality [37] and future interventions [45]. Nevertheless, applications of DNNs on long-sequence (e.g. 12 hours in our data) physiological data are still open, which is one of the problems we aim to tackle in the paper.

[1]https://www.physionet.org/tutorials/hrv/

To the best of our knowledge, few works have been done on integrating the streaming and discrete EHR data. Although the problem of multi-resolution produced in data integrations could be solved by a combination of refining high-density inputs into low-density representations [40] and filling sparse inputs with smoothed missing values [3], advanced methods are still in need for extracting different levels of information from the integrated multimodel data.

**Attention-based Interpretable Deep Models.** Deep learning techniques have often been treated as a black box, but domains like healthcare need both accurate and interpretable results that can assist clinicians in decision making. Attention is one of the most useful mechanisms that launch interoperability into deep neural networks. Extensive attention-based deep models have been developed for images [14, 21, 36, 48], audios and videos [23, 49], machine translations [1, 30, 46]. Attention has also been employed to deep learning models on EHR data. A reverse time attention mechanism was proposed for identifying influential patient hospital visits and medical codes in predictions of heart failure [7]. Later on, graph based attention [6] and hierarchical attention [42] were adapted for incorporating external information like medical ontologies into the deep EHR models; different deep architectures like bidirectional RNN [32], GRU [24] were also applied for improving predictive accuracy. The main idea of these work is using RNN to generate attention weights for mimicking the decision processes of physicians; this is needed because diagnosis in EHR database are not documented in time but are coded together at the end of each visit. The attention mechanism in our model, by contrast, is using multilayer perceptron (MLP) [11, 27] to generate attention weights because our input data are continuous and recorded sequentially given each channel. Our proposed method with guided multi-channel attention is based on prior works [21, 23].

## 3 METHODS

In this section, we describe the components in RAIM: multimodal input processing, guided multi-channel attention and predictive output modeling. Before explaining the details of RAIM, we first introduce the notations.

**Notations.** We denote multimodal EHR data set as $\mathcal{D} = \{C, \mathcal{W}, \mathcal{V}\}$, an integrated set of discrete clinical records $C$, high-density waveforms such as ECG as $\mathcal{W}$ and numerical vital signs such as temperature $\mathcal{V}$. Given an ICU visit indexed by $i$, a clinical record $C_i = \{b_i, R_i, e_i\}$ consist of $b_i$ a set of baseline variables such as demographics; $R_i$ a table of regularly charted variables such as hourly measured heart rate, mean blood pressure; and $e_i$ a list of irregularly recorded events such as lab measurements, drug intakes. Aligned with the $i$th visit, we denote multi-channel waveforms $\mathcal{W}_i = \{f_i^{\text{ch}_n}(t) : 0 \le t \le T_i, \text{ for } n = 1, ..., |\mathcal{W}|\}$, where $T_i$ is the record length for the $i$th visit and $f_i^{\text{ch}_n}$ is the continuously monitored waveform by channel $n$; an example of multi-channel waveforms can be multi-lead ECG recordings sampled at 125Hz. Lastly, we denote multi-channel vital signs $\mathcal{V}_i = \{g_i^{\text{ch}_m}(t) : 0 \le t \le T_i, \text{ for } m = 1, ..., |\mathcal{V}|\}$, where $g^{\text{ch}_m}$ is the $m$th vital sign sampled minutely or secondly; multi-channel vital signs include but not limit to blood pressure, pulse, respiratory rate, SPO2, etc. Table 1 lists all the notations we use in this paper.

Table 1: A list of notations defined in this paper

| Notation | Description |
| --- | --- |
| $\mathcal{D}$ | Multimodal EHR data |
| $C$ | Clinical discrete data |
| $\mathcal{W}$ | Waveform data |
| $\mathcal{V}$ | Vital signs |
| $b$ | Time-invariant baseline variables contained in $C$ |
| R | Regularly charted variables contained in $C$ |
| $e$ | Irregularly measured events contained in $C$ |
| $a$ | Embedded representations from $\mathcal{W}$ and $\mathcal{V}$ |
| $x$ | Time-varying input features derived from $R$ and $e$ |
| $\mathcal{G}$ | Guidance matrix derived from $e$ |

### 3.1 Multimodal Input Processing

For the rest of the paper, we omit subscript $i$ in the above notations and present our method for a single visit in general. Given an observation window $W$ and a time step $t$, we look back at most $T$ steps taking the observed multimodal data $\mathcal{D}_{(\max(t-W,0), t]}$ as input, and predict the output variable $y_t$ at the current step $t$. The output variable can be real or categorical, depending on the task of interest. Decisions about how frequently the output is generated, or equivalently saying how long each time step lasts (i.e., the step length), and how much of data to look back (i.e. the observation window size) also vary by tasks.

Given the step length predefined, we can split raw physiological data $\mathcal{W}$ and $\mathcal{V}$ into a sequence of fixed-length segments indexed by time step $t$. For example, given a step with length of 10 minutes, an ECG segmentation sampled at 125Hz contains in total 75,000 numerical values and a vital sign sampled per second contains 600 values. We posit a CNN on each channel for embedding these high-density segments into low-dimensional representations. We denote the embedded physiological representations as $a_t^{\text{ch}_k}$ for $k = 1, 2, ..., K = |\mathcal{W}| + |\mathcal{V}|$.

For the discrete clinical data $C$, we leave the baseline variables $b$ (e.g. age, gender, etc) as it is. We group the chart table $R$ by rows per time step and calculate the minimum, mean and maximum values per variable given each column; we fill in a missing value with the most recently observed value for the variable. We denote the processed charted variables as an input vector $x_t^{\text{chart}}$. We process the event list $e$, which includes lab and intervention events, into two input structures: $x_t^{\text{lab}}$ an input vector of the real-valued lab measurements and $\mathcal{G}_t$ a guidance matrix derived from the onsets of the lab tests and interventions. Given step $t$, $x_t^{\text{lab}}$ contains the most recently measured value for each lab test and also a binary indicator telling whether the value was newly measured or duplicated.

Given the observation window $W$ and current time step $t$, we define the guidance matrix $\mathcal{G}_t$ as a binary matrix with size of $2 \times \min(t, W)$. We assign $\mathcal{G}_t[0, j] = 1$ if a lab test was newly measured within step $(\max(t - W, 0) + j)$ and 0 otherwise; $\mathcal{G}_t[1, j] = 1$ if an intervention was initiated within the step and 0 otherwise. An intervention initiation can include a new medication administration, IV input initiation, ventilator initiation etc. The intuition behind this matrix construction is to locate potentially important episodes that could influence the final prediction; so attentions can be efficiently concentrated on these episodes rather than spread out all over the observation period. Variations of $\mathcal{G}_t$ can be derived based on different experimental designs or expert knowledge. For example, $\mathcal{G}_t[1, j]$ is assigned as 1 only if a decisive medication was prescribed or a fluid bolus was given. In this paper, RAIM only takes the above generic matrix as the attention guidance; we will leave more refined constructions into future work.

In summary, we have processed the multimodal input data for a single visit into a sequence of embedded step-wise physiological representations $a_t = \{a_t^{\text{ch}_k} : k = 1, ..., K\}$ and $x_t = \{x_t^{\text{chart}}, x_t^{\text{lab}}\}$, binary guidance matrices $\mathcal{G}_t$ and a time-invariant baseline vector $b$. Based on the above inputs, the goal of RAIM is to predict the output variable $y_t$ at time step $t$.

## 3.2 Guided Multi-Channel Attention

Given the observation window $W$ and a sequence of embedded physiological inputs $a_{(\max(t-W,0),t]}$ up to (including) time $t$, we use RNN to encode the sequential data. More specifically, RNN generates hidden states $h_\tau$ based on the history $a_{(\max(t-W,0),\tau]}$ for any $\tau \in (\max(t - W, 0), t]$. The final state $h_t$ will be used for predicting the output $y_t$ at the end. Below we introduce two attention mechanisms developed in the model: multi-channel attention for identifying which input channel influence most on the final output prediction; guidance-based attention for effectively identifying which episodes (i.e. time steps) influence most on the final prediction. Without loss of generality, we write the input sequence as $a_{(t-W,t]}$ for notation simplification. This assumes $t \geq W$ and the sequence length is $W$; alternatively if $t < W$, the notation becomes $a_{(0,t]}$ and the length becomes $t$.

### 3.2.1 Multi-channel Attention.
We break down the input sequence $a_{(t-W,t]}$ by time and obtain time-specific input vectors $a_\tau = a_\tau^{\text{ch}_1} \oplus ... \oplus a_\tau^{\text{ch}_K}$ for $\tau \in (t - W, t]$, where $\oplus$ denotes the vector operation of concatenation. Similarly, we break down the inputs by channels and obtain channel-specific input vectors $a^{\text{ch}_k} = a_{t-W+1}^{\text{ch}_k} \oplus ... \oplus a_t^{\text{ch}_k}$ for $k \in [1, K]$. The goal of the multi-channel attention is to learn two vectors of weights: a $1 \times W$ time-specific weight vector $\alpha_t$ and a $1 \times K$ channel-specific weight vector $\beta_t$. By taking outer product of the two weight vectors, we obtain a $K \times W$ weight matrix $A_t = \beta_t^\mathsf{T} \alpha_t$. Figure 2 describes the idea, where $h_{t-1}$ is the encoded history state prior to time $t$ and $Z_t$ is a convex combination of the re-weighted multi-channel inputs. More

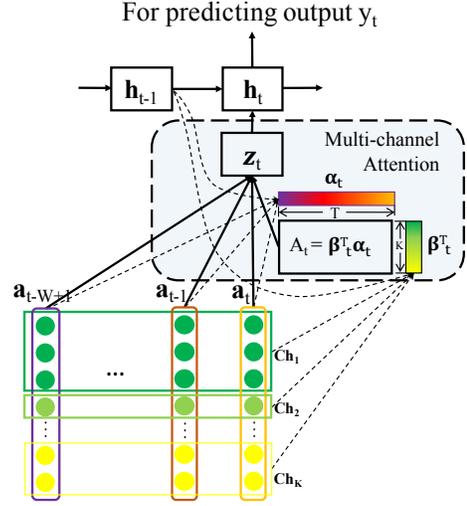

**Figure 2: Multi-channel attention in RAIM**

specifically, $Z_t$ is computed as follows

$$Z_t = \sum_{j=1}^{W} \alpha_{tj}(\beta_t * a_{t-W+j}), \quad (1)$$

where $\beta_t * a_\tau = \beta_{t1} a_\tau^{\text{ch}_1} \oplus ... \oplus \beta_{tK} a_\tau^{\text{ch}_K}$ for $\tau \in (t - W, t]$.

To obtain vector $\alpha_t$, we train a multilayer perceptron (MLP) on hidden state $h_{t-1}$ and time-specific input vectors $a_\tau$ for $\tau \in (t - W, t]$, to generate a $1 \times T$ energy vector $s_t^{\text{time}}$:

$$s_t^{\text{time}} = \tanh(W_h^\alpha \cdot h_{t-1} + a_{(t-W,t]}^\mathsf{T} w_a^\alpha + b^\alpha), \quad (2)$$

where $W_h^\alpha \in \mathbb{R}^{W \times |h|}$ and $w_a^\alpha \in \mathbb{R}^{|a| \times 1}$ are weighted matrix and vector, $b^\alpha \in \mathbb{R}^{W \times 1}$ is the bias vector; $|h|$ and $|a|$ are the dimensions of $h_{t-1}$ and $a_t$ respectively. We obtain the final attention weights using a Softmax function:

$$\alpha_{tj} = \frac{\exp(s_{tj}^{\text{time}})}{\sum_{j'=1}^{W} \exp(s_{tj'}^{\text{time}})}, \text{ for } j = 1, ..., W. \quad (3)$$

Similarly, we get vector $\beta_t$ by training another MLP on $h_{t-1}$ and channel-specific input vectors $a^{\text{ch}_k}$ for $k \in [1, K]$:

$$s_t^{\text{ch}} = \tanh(W_h^\beta \cdot h_{t-1} + a^{\text{ch}_1:\text{ch}_K} w_a^\beta + b^\beta), \quad (4)$$

$$\beta_{tk} = \frac{\exp(s_{tk}^{\text{ch}})}{\sum_{k'=1}^{K} \exp(s_{tk'}^{\text{ch}})}, \text{ for } k = 1, ..., K,$$

where $W_h^\beta$, $w_a^\beta$, and $b^\beta$ are the parameters learnt in MLP.

### 3.2.2 Guidance-based Attention.
Guidance from external knowledge can be useful for effectively generating attention weights; only a subset of the $W$ episodes need to be attended so the latent unit $Z_t$ can also be computed efficiently. Given the guidance matrix $\mathcal{G}$ we derived in Section 3.1, we locate two types of episodes that can potentially change the final output: the episodes when one or more lab tests were conducted and the episodes when an intervention was initiated. Below we describe these two guided attention mechanisms in details.

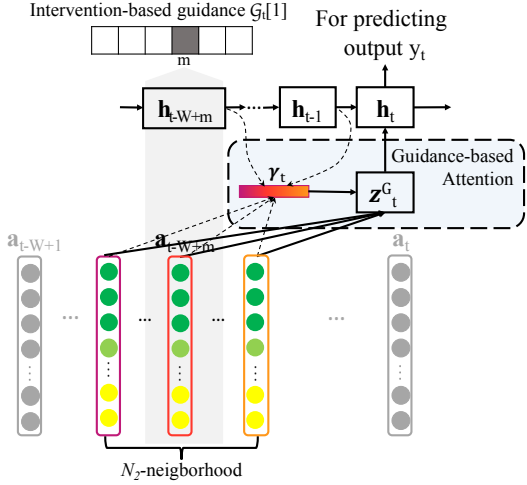

**Figure 3: An example of guided attention derived from intervention events in RAIM**

**Lab-measurement guided attention.** Given the guidance row vector $\mathcal{G}_t[0]$ derived from lab events, we assume that the $N_1$-neighbors around step $t - W + j$ will be attended if and only if $\mathcal{G}_t[0, j] = 1$; we call these steps as 'active' steps. Thus we only collect the input vectors at active steps, fit them with the encoded state $h_{t-1}$ into MLP, and obtain a smaller set of active attention weights. The hidden unit $Z_t^{\mathcal{G}}$ is computed as

$$Z_t^{\mathcal{G}} = \sum_{j \in \Phi_0} \gamma_{tj} a_{t-W+j}, \qquad (5)$$

where $\Phi_0 = \{j : \exists j' \text{ s.t. } \mathcal{G}_t[0, j'] = 1 \text{ and } |j - j'| \leq N_1/2\}$ denotes the set of active time steps and $\gamma_t$ is the shortened weight vector. The generation of $\gamma_t$ follows equations (2) and (3), whereas $a_{(t-W, t]}$ the entire input sequence is substituted by $a_{[j : j \in \Phi_0]}$ the active inputs derived by guidance $\mathcal{G}[0]$.

**Intervention guided attention.** Given the row vector $\mathcal{G}_t[1]$ derived from intervention events, we similarly assume that the $N_2$-neighbors around step $t - W + j$ will be attended if and only if $\mathcal{G}_t[1, j] = 1$. Besides, we extract the encoded state $h_{t-W+m}$ at the most recently intervened step $t - W + m$, i.e., $m = \max\{j : \mathcal{G}_t[1, j] = 1\}$. Different from the previous formulation, we send two encoded states $h_{t-T+m}$ and $h_{t-1}$ together with the active input vectors into MLP; the motivation is to cast the relationship between the two states into weight generations, rather than using a single state $h_{t-1}$. The relationship can be interpreted as whether the interventions have effects on changing the hidden state of the patient. Figure 3 illustrates the idea. The hidden unit $Z_t^{\mathcal{G}}$ is computed the same as in Eq. (5) by substituting $\Phi_0$ with $\Phi_1$, where $\Phi_1 = \{j : \exists j' \text{ s.t. } \mathcal{G}_t[1, j'] = 1 \text{ and } |j - j'| \leq N_2/2\}$. The weight vector $\gamma_t$ is generated by MLP similarly as before with additional parameter corresponding to $h_{t-W+m}$.

### 3.3 Predictive Output Modeling

For sequentially encoding the physiological history $h_t$'s, we use a standard configuration of LSTM [17]. Given the encoded state $h_t$, the time-varying variable $x_t$ and the time-invariant baseline variable $b$, we can predict a categorical output variable $y_t$ using multivariate regression

$$\hat{y}_t = \text{Softmax}(W_h^y h_t + W_x^y x_t + b^\top w^y + b^y), \qquad (6)$$

where $W_h^y$, $W_x^y$, $w^y$ and $b^y$ are the parameters to be learned. Given $I$ independent visits observed in the data, we use cross-entropy to calculate the final loss:

$$\mathcal{L} = -\sum_{i=1}^{I} \sum_{t=1}^{T_i} \Big( y_{it} \log(\hat{y}_{it}) + (1 - y_{it}) \log(1 - \hat{y}_{it}) \Big). \qquad (7)$$

Here we get back the subscript $i$ and let $\hat{y}_{it}$ denote the predicted risk score for the $i$th visit at step $t$. Our model is trained end-to-end using backpropagation. Alternatively for predicting a real-valued output variable $y_t$, Eq. (6) can be simplified as a linear regression and mean squared error will be used for computing the final loss.

## 4 EXPERIMENTS

We evaluate our model RAIM in two prediction tasks based on a publicly available real data set. We show that RAIM outperforms our baselines in both quantitative analysis and qualitative analysis.

### 4.1 Data

**Data set**. We demonstrate our model on the MIMIC-III Waveform Database Matched Subset [2] [22], a publicly available multimodal EHR data released last year on PhysioNet [15]. The data integrates deidentified and comprehensive clinical data with the continuously monitored physiological data from bedside monitors in adult and neonatal intensive care units (ICU) at the Beth Israel Deaconess Medical Center in Boston. This matched subset contains 22,317 waveform records (including multi-leads ECG signals, fingertip photoplethysmogram (PPG) signals, and up to 8 waveforms simultaneously), 22,247 numerics records (time series of vital signs sampled per minute) and 10,282 clinical discrete records. We evaluate RAIM in two prediction tasks[3] based on the data.

**Prediction task 1**. The first task is to detect physiologic *decompensation* in an ICU visit. Monitoring in ICUs are often equipped with early warning scores or alarm generations; a useful way of evaluating such scores is to accurately predict mortality within a fixed time window [47]. In this experiment, we formulate the detection problem as a binary classification problem, that is to predict whether a patient will die within the next 24 hours.

**Prediction task 2**. The second task is to forecast the *length of stay (LOS)* at ICU. Patients with longer LOS often indicate they have more severe and complex conditions, and require more hospital resources and costs. In this experiment, we divide LOS into 9 buckets and focus on a multiclass classification problem. For example, class 1–7 correspond to 1–7 days of stay respectively, class 8 corresponds to more than 8 days and up to two weeks of stay, and the final class 9 corresponds to over two weeks of stay.

**Data processing**. We define the observation window $W = 12$ hours in our experiments, and process the same input data for the two prediction tasks. We first use a toolkit [4] to generate the two labels at the end of the observation window and extract the clinical features: baseline variables $b$ include *age, gender*, and *ethnicity*, hourly charted variables $x_t^{\text{chart}}$'s include *oxygen saturation, diastolic/systolic/mean BP, heart rate* and *respiratory rate*, and irregular lab measurements $x_t^{\text{lab}}$'s include *glucose, PH*, and *temperature*. To generate the guidance matrix $\mathcal{G}$, we consider the above three lab events and intervention events including *procedures* and *IV inputs*. Lastly, we extract the waveform $\mathcal{W}$ as the 125Hz *lead-II ECG waveform*, and the vital signs $\mathcal{V}$ including minutely sampled *systolic ABP, diastolic ABP, systolic NBP, diastolic NBP, Pulse, Respiration rate* and *SpO2*. Our cohort include only adult patients (age older than 18 at the time of ICU admission) whose ICU stays are beyond 13 hours since their first ECG recordings are available in the database. This results at a cohort including 10, 988 ICU visits from 6, 670 unique patients with 11.4% mortality. Figure 4 shows the detailed statistics of the demographics and LOS in this cohort. For the sake of data quality, we discard the first 1 hour of the data since equipments were still possibly set up. Given $W = 12$ hours, we segment the sequences into 12-hour long time series. As a result, we obtain in total 32, 868 time series, 27, 762 (from 85% of the total patients) out of which are put to the training set and the rest to the test set.

---
[2]https://physionet.org/physiobank/database/mimic3wdb/matched/
[3]The two prediction tasks were originally defined in [16], a recently public multitask benchmark on MIMIC-III Clinical Database; but the data were not matched with the MIMIC-III Waveform Database.
[4]https://github.com/YerevaNN/mimic3-benchmarks

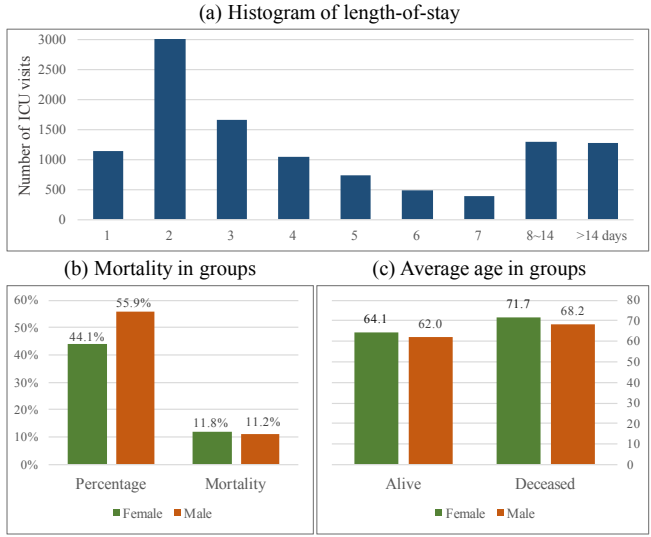

**Figure 4: Statistics of our cohort for predicting length-of-stay and decompensation of patients in ICU**

## 4.2 Experimental setup

**Model configurations and Training details**. We define the step size in RNN to be 1 hour, and train a RNN of length 12 given the observation window $W = 12$ hours. As a result, the ECG signal within a time step contains up to 450, 000 numerical values, and the minutely sampled vital signs in each channel contains 60 values. In this experiment, we only posit CNN on the ECG signals to obtain 128-dimensional embedded representations; we use the raw 60 values for each of the 7 vital signs but would recommend to posit CNN again on each signal if they were sampled more densely like secondly. Prior work [40] has shown CNN can perform well in classifying 30-second ECG signals; for embedding the 1-hour ECG signals in our experiment, we initially trained CNN on 30-second segmentations and train another CNN or RNN on top of the 1-hour concatenated outputs, but these didn't outperform a single CNN as we report here in this paper. For training the CNN, we explore different number of layers that gradually increases from 3, 5, 11 to 34. We use *batch normalization*, *ReLU* activation and *max pooling* in between convolutional layers [18], and use *SAME padding* in the model. For the final model, we reach at 5-layer CNN with kernel size varying from 10 to 3 as it goes deeper.

For the RNN predictive model, we use 3-layer bi-directional LSTM and explore different configurations in the model: activation functions include *tanh*, *ReLU*, and *PReLU*; optimizers include ADAM, ADAGRAD, Stochastic Gradient Descent and RMSProp; batch size varies between 32, 64, and 128. We use random search [2] to find the optimal model at the end, which uses ADAM optimizer, *ReLU* activation and batch size of 32.

**Implementation details.** We implement all the models with PyTorch 0.3.0 [38]. For training models, We use Adam [25] with a batch size of 32 samples on a machine equipped with Intel Xeon E5-2640, 256GB RAM, eight Nvidia Titan-X GPU and CUDA 8.0.

**Baselines**. Here we list the models we compare in our experiments.

- **CNN (ECG)**: The CNN model trained on 1-hour ECG signals.
- **CNN-RNN**: The vanilla CNN-RNN model trained on the full 12-hour time series. Rest models will be trained on sequential data too.
- **CNN-AttRNN**: The conventional attention model on RNN (i.e., no channel-specific attentions).
- **CNN-MultiChAttRNN (RAIM-0)**: Initial version of RAIM with only multi-channel attention.
- **CNN-LabMultiChAttRNN (RAIM-1)**: Next version of RAIM with multi-channel attention guided by lab measurements.
- **CNN-IntMultiChAttRNN (RAIM-2)**: Alternative version of RAIM with multi-channel attention guided by interventions.
- **CNN-IntLabMultiChAttRNN (RAIM-3)**: Final version of RAIM with multi-channel attention guided by both lab and interventions (The two learned $Z_t^{\mathcal{G}}$ vectors are concatenated as one input vector for generating $h_t$).

## 4.3 Results

We report both quantitative and qualitative results from our experiments. In quantitative results, we compare prediction performance of the 7 models on the two tasks. In qualitative results, we illustrate the effectiveness of RAIM's attention mechanism by showing the meaningful active time steps influencing the risk prediction of decompensation in a new test patient. We also present tSNE [33] plots generated from the encoded states learnt by LSTM and show the final RAIM model obtain better representations w.r.t. predicting length of stay.

*4.3.1 Quantitative results.* For the binary classification task of predicting decompensation, we evaluate our models in terms of AUC-ROC, AUC-PR and Accuracy (cutoff of 0.5). For the multiclass prediction task of forecasting length of stay, we evaluate the models in terms of Cohen's Kappa [10] and Accuracy. The kappa score ranges between −1 and 1, and scores above .8 are usually considered as good agreement.

Table 2: Performance comparison of the 7 models on predicting decompensation and length of stay

|  | Decompensation | | | Length of Stay | |
| --- | --- | --- | --- | --- | --- |
|  | AUC-ROC | AUC-PR | Accuracy | Kappa | Accuracy |
| CNN (ECG) | 87.84% | 21.56% | 88.38% | 0.7681 | 82.16% |
| CNN-RNN | 87.45% | 23.19% | 88.25% | 0.8027 | 85.34% |
| CNN-AttRNN | 88.19% | 25.81% | 89.28% | 0.8186 | 84.89% |
| RAIM-0 | 87.81% | 25.56% | 88.96% | 0.8125 | 85.84% |
| RAIM-1 | 88.25% | 25.61% | 88.91% | 0.8215 | 86.74% |
| RAIM-2 | 88.77% | 26.85% | 90.27% | 0.8217 | 85.21% |
| RAIM-3 | **90.18%** | **27.93%** | **90.89%** | **0.8291** | **86.82%** |

Table 2 reports the prediction scores on the two prediction tasks. We see that RAIM-3 outperforms all other models in both tasks, obtaining an AUC-ROC score of 90.18%, AUC-PR of 27.93%, Accuracy of 90.89% for predicting decompensation and a Kappa of 0.8291 and Accuracy of 86.82% for predicting length of stay. CNN (ECG) has a good performance in predicting physiological decompensation but low performance in predicting LOS; it indicates these high-dense signals are rich enough for some of the prediction tasks but also need integrated with other data modalities for complex tasks like prediting LOS. CNN-AttRNN has higher scores compared to CNN-RNN model indicating that the attention mechanism helps in performance improvement. Adding muti-channel attention (RAIM-0) also improves CNN-RNN, but didn't beat CNN-AttRNN; however it provides additional interpretability into the model. Multi-channel attention guided by lab or intervention events (RAIM-1 and RAIM-2) starts to outperform the previous models. Lastly, the final model RAIM-3 incorporating guidance from both lab and intervention events reaches the best performance on both tasks.

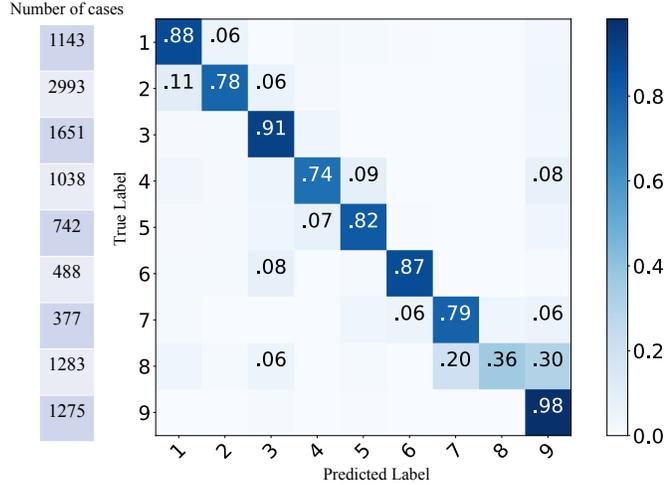

Figure 5: Confusion matrix for predicting the 9 classes of length-of-stay by RAIM-3: class 1-7 corresponds to 1-7 days, class 8 corresponds to 7-14 days, and class 9 corresponds to over 14 days

In addition, we plot the confusion matrix for RAIM-3 predicting the 9 classes of LOS in Figure 5. It is shown that predicting patients' stays to be more than 1 week but less than 2 weeks (class 8) is the hardest task; it is confused with the lower class and more confused with the extreme class that patients may stay over two weeks.

*4.3.2 Qualitative results.* To demonstrate the effectiveness of the proposed attention mechanism in RAIM, we plot the real-time risk prediction for detecting physiological deterioration on an unseen test patient in Figure 6. The patient deceased in 13 hours after the admission to ICU. The predicted risk score increases from 0.49 to 0.72 in the 12-hour observation window. The patient is predicted having higher risk of decompensation than average at the 4th hour when high attentions (highlighted as orange) are generated on multiple channels from RAIM; this matches the observation that clinicians decided to initiate the first intervention at the same time. Another high attentions across multiple channels are produced 2 hours prior to the patient's death.

In Figure 7, we randomly select 3, 000 time series from the test set and plot their encoded representations $h_t$'s at the final step in the prediction task of LOS. Each colored dot in the tSNE plot represents a test case; the 9 colors correspond to the 9 classes of LOS. We show three tSNE plots generated from three different models: CNN-RNN, RAIM-1, RAIM-3. We observe that the tSNE embeddings of the final representation learnt by LSTM are better distinguishable in RAIM-0 and RAIM-3, where RAIM-3 shows the best. This leads to the conclusion that efficiently guided attention model helps in obtaining better representations.

## 5 CONCLUSIONS

High-density multi-channel signals such as ECG signals integrated with discrete clinical records can be very useful in risk modeling in ICU patients. we propose RAIM to overcome the challenges associated with modeling high-density multimodal data and incorporating interpretability and efficiency into the modeling.

We conduct a thorough evaluation of RAIM using two important clinical prediction tasks: predicting length of stay and physiological decompensation. We assess different variations of RAIM model and show the prediction performance is improved as we develop more comprehensive attention mechanisms in the model. Qualitative analysis of showing the meaningful attended channels and episodes in risk score prediction provide evidence of interpretablity in RAIM.

Future work includes extracting more refined or task-specific guidance matrix based on domain knowledge, predicting other crucial outcomes in critically ill patients such as cardiac arrest, and testing on other data modalities.

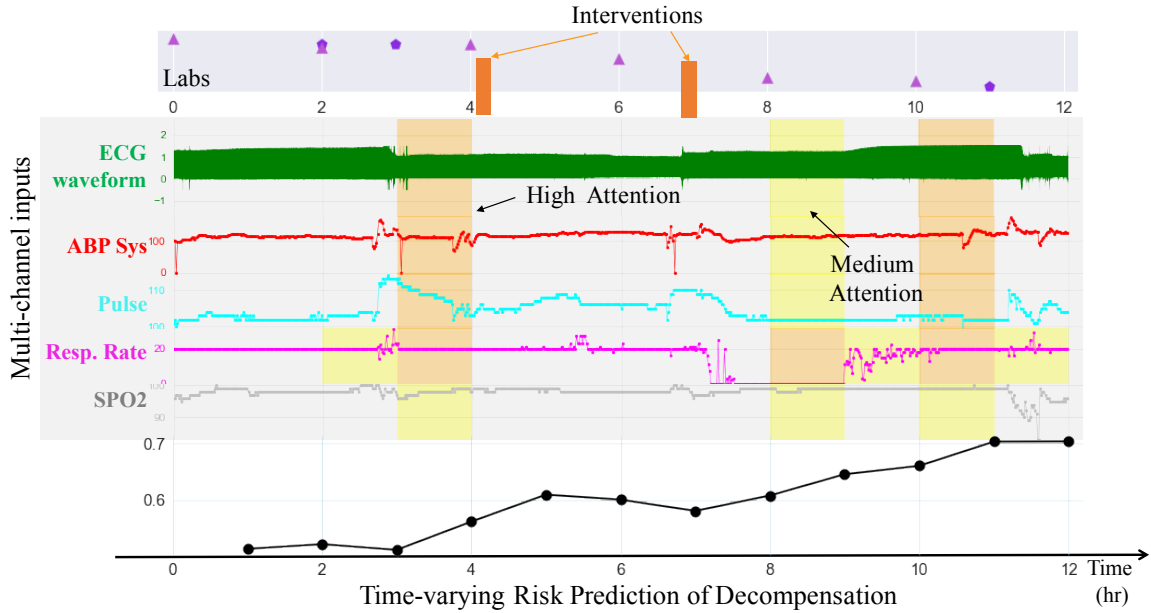

**Figure 6: Time-varying risk prediction of decompensation by `RAIM-3` on an unseen test patient who deceased at the 13rd hour. The learnt attention regions are highlighted in yellow (low element-wise weights between 0.01 to 0.02) and orange (high element-wise weights between 0.02 to 0.07).**

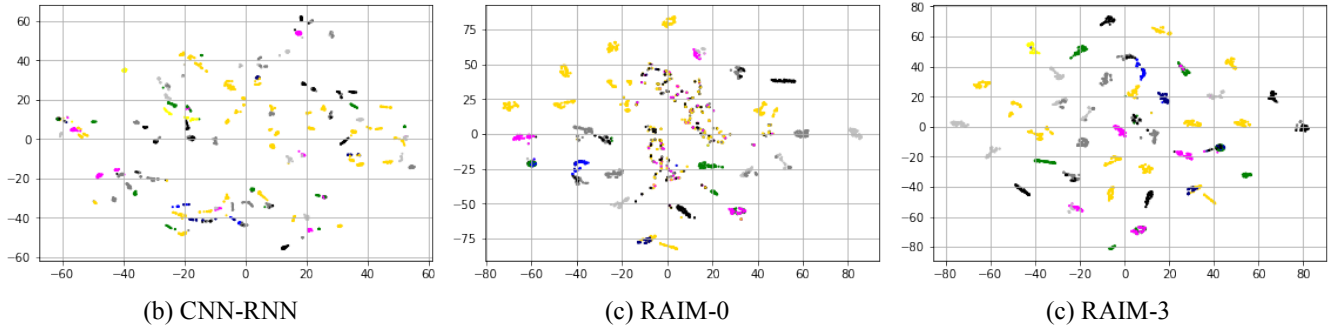

**Figure 7: The tSNE plots of the final representations learnt from LSTM in CNN-RNN, `RAIM-0`, and `RAIM-3` for the multiclass prediction task of forecasting length of stay. Representations become more distinguishable as we improve the baseline model.**

## ACKNOWLEDGMENTS
This work was supported by the National Science Foundation, award IIS-#1418511 and CCF-#1533768, the National Institute of Health award 1R01MD011682-01 and R56HL138415, and Children's Healthcare of Atlanta.

## REFERENCES
[1] Dzmitry Bahdanau, Kyunghyun Cho, and Yoshua Bengio. 2014. Neural machine translation by jointly learning to align and translate. *arXiv preprint arXiv:1409.0473* (2014).
[2] James Bergstra and Yoshua Bengio. 2012. Random search for hyper-parameter optimization. *Journal of Machine Learning Research* 13, Feb (2012), 281–305.
[3] Zhengping Che, Sanjay Purushotham, Kyunghyun Cho, David Sontag, and Yan Liu. 2016. Recurrent neural networks for multivariate time series with missing values. *arXiv preprint arXiv:1606.01865* (2016).
[4] Edward Choi, Mohammad Taha Bahadori, Andy Schuetz, Walter F Stewart, and Jimeng Sun. 2016. Doctor ai: Predicting clinical events via recurrent neural networks. In *Machine Learning for Healthcare Conference*. 301–318.
[5] Edward Choi, Mohammad Taha Bahadori, Elizabeth Searles, Catherine Coffey, and Jimeng Sun. 2016. Multi-layer Representation Learning for Medical Concepts. *arXiv preprint arXiv:1602.05568* (2016).
[6] Edward Choi, Mohammad Taha Bahadori, Le Song, Walter F Stewart, and Jimeng Sun. 2017. GRAM: Graph-based attention model for healthcare representation learning. In *Proceedings of the 23rd ACM SIGKDD International Conference on Knowledge Discovery and Data Mining*. ACM, 787–795.
[7] Edward Choi, Mohammad Taha Bahadori, Jimeng Sun, Joshua Kulas, Andy Schuetz, and Walter Stewart. 2016. Retain: An interpretable predictive model for healthcare using reverse time attention mechanism. In *Advances in Neural Information Processing Systems*. 3504–3512.
[8] Edward Choi, Andy Schuetz, Walter F Stewart, and Jimeng Sun. 2016. Medical concept representation learning from electronic health records and its application on heart failure prediction. *arXiv preprint arXiv:1602.03686* (2016).
[9] Edward Choi, Andy Schuetz, Walter F Stewart, and Jimeng Sun. 2016. Using recurrent neural network models for early detection of heart failure onset. *Journal*

*of the American Medical Informatics Association* 24, 2 (2016), 361–370.
[10] Jacob Cohen. 1960. A coefficient of agreement for nominal scales. *Educational and psychological measurement* 20, 1 (1960), 37–46.
[11] Dumitru Erhan, Yoshua Bengio, Aaron Courville, and Pascal Vincent. 2009. Visualizing higher-layer features of a deep network. *University of Montreal* 1341, 3 (2009), 1.
[12] Cristóbal Esteban, Oliver Staeck, Stephan Baier, Yinchong Yang, and Volker Tresp. 2016. Predicting clinical events by combining static and dynamic information using recurrent neural networks. In *Healthcare Informatics (ICHI), 2016 IEEE International Conference on*. IEEE, 93–101.
[13] Jason Alan Fries. 2016. Brundlefly at SemEval-2016 Task 12: Recurrent neural networks vs. joint inference for clinical temporal information extraction. *arXiv preprint arXiv:1606.01433* (2016).
[14] Jianlong Fu, Heliang Zheng, and Tao Mei. 2017. Look closer to see better: Recurrent attention convolutional neural network for fine-grained image recognition. In *Conf. on Computer Vision and Pattern Recognition*.
[15] Ary L Goldberger, Luis AN Amaral, Leon Glass, Jeffrey M Hausdorff, Plamen Ch Ivanov, Roger G Mark, Joseph E Mietus, George B Moody, Chung-Kang Peng, and H Eugene Stanley. 2000. Physiobank, physiotoolkit, and physionet. *Circulation* 101, 23 (2000), e215–e220.
[16] Hrayr Harutyunyan, Hrant Khachatrian, David C Kale, and Aram Galstyan. 2017. Multitask Learning and Benchmarking with Clinical Time Series Data. *arXiv preprint arXiv:1703.07771* (2017).
[17] Sepp Hochreiter and Jürgen Schmidhuber. 1997. Long short-term memory. *Neural computation* 9, 8 (1997), 1735–1780.
[18] Sergey Ioffe and Christian Szegedy. 2015. Batch normalization: Accelerating deep network training by reducing internal covariate shift. *arXiv preprint arXiv:1502.03167* (2015).
[19] Abhyuday N Jagannatha and Hong Yu. 2016. Bidirectional RNN for medical event detection in electronic health records. In *Proceedings of the conference. Association for Computational Linguistics. North American Chapter. Meeting*, Vol. 2016. NIH Public Access, 473.
[20] Abhyuday N Jagannatha and Hong Yu. 2016. Structured prediction models for RNN based sequence labeling in clinical text. In *Proceedings of the Conference on Empirical Methods in Natural Language Processing. Conference on Empirical Methods in Natural Language Processing*, Vol. 2016. NIH Public Access, 856.
[21] Xu Jia, Efstratios Gavves, Basura Fernando, and Tinne Tuytelaars. 2015. Guiding the long-short term memory model for image caption generation. In *Computer Vision (ICCV), 2015 IEEE International Conference on*. IEEE, 2407–2415.
[22] Alistair EW Johnson, Tom J Pollard, Lu Shen, H Lehman Li-wei, Mengling Feng, Mohammad Ghassemi, Benjamin Moody, Peter Szolovits, Leo Anthony Celi, and Roger G Mark. 2016. MIMIC-III, a freely accessible critical care database. *Scientific data* 3 (2016), 160035.
[23] Suyoun Kim and Ian Lane. 2015. Recurrent models for auditory attention in multi-microphone distance speech recognition. *arXiv preprint arXiv:1511.06407* (2015).
[24] You Jin Kim, Yun-Geun Lee, Jeong Whun Kim, Jin Joo Park, Borim Ryu, and Jung-Woo Ha. 2017. Highrisk Prediction from Electronic Medical Records via Deep Attention Networks. *arXiv preprint arXiv:1712.00010* (2017).
[25] Diederik Kingma and Jimmy Ba. 2014. Adam: A method for stochastic optimization. *arXiv preprint arXiv:1412.6980* (2014).
[26] Thomas A Lasko, Joshua C Denny, and Mia A Levy. 2013. Computational phenotype discovery using unsupervised feature learning over noisy, sparse, and irregular clinical data. *PloS one* 8, 6 (2013), e66341.
[27] Quoc V Le. 2013. Building high-level features using large scale unsupervised learning. In *Acoustics, Speech and Signal Processing (ICASSP), 2013 IEEE International Conference on*. IEEE, 8595–8598.
[28] Zachary C Lipton, David C Kale, Charles Elkan, and Randall Wetzel. 2015. Learning to diagnose with LSTM recurrent neural networks. *arXiv preprint arXiv:1511.03677* (2015).
[29] Yue Liu, Tao Ge, Kusum Mathews, Heng Ji, and Deborah McGuinness. 2015. Exploiting task-oriented resources to learn word embeddings for clinical abbreviation expansion. *Proceedings of BioNLP 15* (2015), 92–97.
[30] Minh-Thang Luong, Hieu Pham, and Christopher D Manning. 2015. Effective approaches to attention-based neural machine translation. *arXiv preprint arXiv:1508.04025* (2015).
[31] Xinbo Lv, Yi Guan, Jinfeng Yang, and Jiawei Wu. 2016. Clinical relation extraction with deep learning. *International Journal of Hybrid Information Technology* 9, 7 (2016), 237–248.
[32] Fenglong Ma, Radha Chitta, Jing Zhou, Quanzeng You, Tong Sun, and Jing Gao. 2017. Dipole: Diagnosis prediction in healthcare via attention-based bidirectional recurrent neural networks. In *Proceedings of the 23rd ACM SIGKDD International Conference on Knowledge Discovery and Data Mining*. ACM, 1903–1911.
[33] Laurens van der Maaten and Geoffrey Hinton. 2008. Visualizing data using t-SNE. *Journal of machine learning research* 9, Nov (2008), 2579–2605.
[34] Saaed Mehrabi, Sunghwan Sohn, Dingcheng Li, Joshua J Pankratz, Terry Therneau, Jennifer L St Sauver, Hongfang Liu, and Mathew Palakal. 2015. Temporal pattern and association discovery of diagnosis codes using deep learning. In *Healthcare Informatics (ICHI), 2015 International Conference on*. IEEE, 408–416.
[35] Riccardo Miotto, Li Li, Brian A Kidd, and Joel T Dudley. 2016. Deep patient: an unsupervised representation to predict the future of patients from the electronic health records. *Scientific reports* 6 (2016), 26094.
[36] Volodymyr Mnih, Nicolas Heess, Alex Graves, et al. 2014. Recurrent models of visual attention. In *Advances in neural information processing systems*. 2204–2212.
[37] Phuoc Nguyen, Truyen Tran, and Svetha Venkatesh. 2017. Deep Learning to Attend to Risk in ICU. *arXiv preprint arXiv:1707.05010* (2017).
[38] Adam Paszke, Sam Gross, Soumith Chintala, Gregory Chanan, Edward Yang, Zachary DeVito, Zeming Lin, Alban Desmaison, Luca Antiga, and Adam Lerer. 2017. Automatic differentiation in PyTorch. (2017).
[39] Trang Pham, Truyen Tran, Dinh Phung, and Svetha Venkatesh. 2016. Deepcare: A deep dynamic memory model for predictive medicine. In *Pacific-Asia Conference on Knowledge Discovery and Data Mining*. Springer, 30–41.
[40] Pranav Rajpurkar, Awni Y Hannun, Masoumeh Haghpanahi, Codie Bourn, and Andrew Y Ng. 2017. Cardiologist-level arrhythmia detection with convolutional neural networks. *arXiv preprint arXiv:1707.01836* (2017).
[41] Patrick Schwab, Gaetano Scebba, Jia Zhang, Marco Delai, and Walter Karlen. 2017. Beat by Beat: Classifying Cardiac Arrhythmias with Recurrent Neural Networks. *arXiv preprint arXiv:1710.06319* (2017).
[42] Ying Sha and May D Wang. 2017. Interpretable Predictions of Clinical Outcomes with An Attention-based Recurrent Neural Network. In *Proceedings of the 8th ACM International Conference on Bioinformatics, Computational Biology, and Health Informatics*. ACM, 233–240.
[43] Benjamin Shickel, Patrick James Tighe, Azra Bihorac, and Parisa Rashidi. 2017. Deep EHR: A Survey of Recent Advances in Deep Learning Techniques for Electronic Health Record (EHR) Analysis. *IEEE Journal of Biomedical and Health Informatics* (2017).
[44] Yan Sun, Kap Luk Chan, and Shankar Muthu Krishnan. 2005. Characteristic wave detection in ECG signal using morphological transform. *BMC cardiovascular disorders* 5, 1 (2005), 28.
[45] Harini Suresh, Nathan Hunt, Alistair Johnson, Leo Anthony Celi, Peter Szolovits, and Marzyeh Ghassemi. 2017. Clinical Intervention Prediction and Understanding with Deep Neural Networks. In *Machine Learning for Healthcare Conference*. 322–337.
[46] Ashish Vaswani, Noam Shazeer, Niki Parmar, Jakob Uszkoreit, Llion Jones, Aidan N Gomez, Łukasz Kaiser, and Illia Polosukhin. 2017. Attention is all you need. In *Advances in Neural Information Processing Systems*. 6000–6010.
[47] B Williams, G Alberti, C Ball, D Bell, R Binks, L Durham, et al. 2012. National early warning score (NEWS): Standardising the assessment of acute-illness severity in the NHS. *London: The Royal College of Physicians* (2012).
[48] Kelvin Xu, Jimmy Ba, Ryan Kiros, Kyunghyun Cho, Aaron Courville, Ruslan Salakhudinov, Rich Zemel, and Yoshua Bengio. 2015. Show, attend and tell: Neural image caption generation with visual attention. In *International Conference on Machine Learning*. 2048–2057.
[49] Bolei Zhou, Alex Andonian, and Antonio Torralba. 2017. Temporal Relational Reasoning in Videos. *arXiv preprint arXiv:1711.08496* (2017).